\pdfoutput=1
\documentclass[11pt]{article}
\usepackage{acl}

\usepackage{times}
\usepackage{latexsym}

\usepackage[T1]{fontenc}

\usepackage[utf8]{inputenc}

\usepackage{microtype}

\usepackage{inconsolata}

\usepackage{tcolorbox}
\usepackage{amsmath}
\usepackage{outlines}
\usepackage{hyperref}
\usepackage{url}
\usepackage{tikz}
\usepackage{bm}
\usepackage{amsmath,amsfonts,amssymb}
\usepackage{float}
\usepackage{listings}
\usepackage{color}
\usepackage{titlesec}
\usepackage{amsthm}
\usepackage{graphicx}
\usepackage{caption}
\usepackage{subcaption}
\usepackage{bbm}
\usepackage[whole]{bxcjkjatype}

\title{Exploring Intra and Inter-language Consistency in Embeddings with ICA}

 \author{Rongzhi Li$^1$, \ Takeru Matsuda$^{1,2}$, \ Hitomi Yanaka$^1$ \\
    $^1$The University of Tokyo,\hspace{0.3cm}$^2$RIKEN Center for Brain Science\\
    \texttt{\{iimori-eiji,hyanaka\}@is.s.u-tokyo.ac.jp} \\
    \texttt{matsuda@mist.i.u-tokyo.ac.jp} \\}

\begin{document}
\maketitle
\begin{abstract}
Word embeddings represent words as multidimensional real vectors, facilitating data analysis and processing, but are often challenging to interpret. 
Independent Component Analysis (ICA) creates clearer semantic axes by identifying independent key features. 
Previous research has shown ICA's potential to reveal universal semantic axes across languages. 
However, it lacked verification of the consistency of independent components within and across languages. 
We investigated the consistency of semantic axes in two ways: both within a single language and across multiple languages. 
We first probed into intra-language consistency, focusing on the reproducibility of axes by performing ICA multiple times and clustering the outcomes.
Then, we statistically examined inter-language consistency by verifying those axes' correspondences using statistical tests. 
We newly applied statistical methods to establish a robust framework that ensures the reliability and universality of semantic axes.

\end{abstract}

\begin{figure}[ht]
  \centering
    \includegraphics[width=0.8\linewidth]{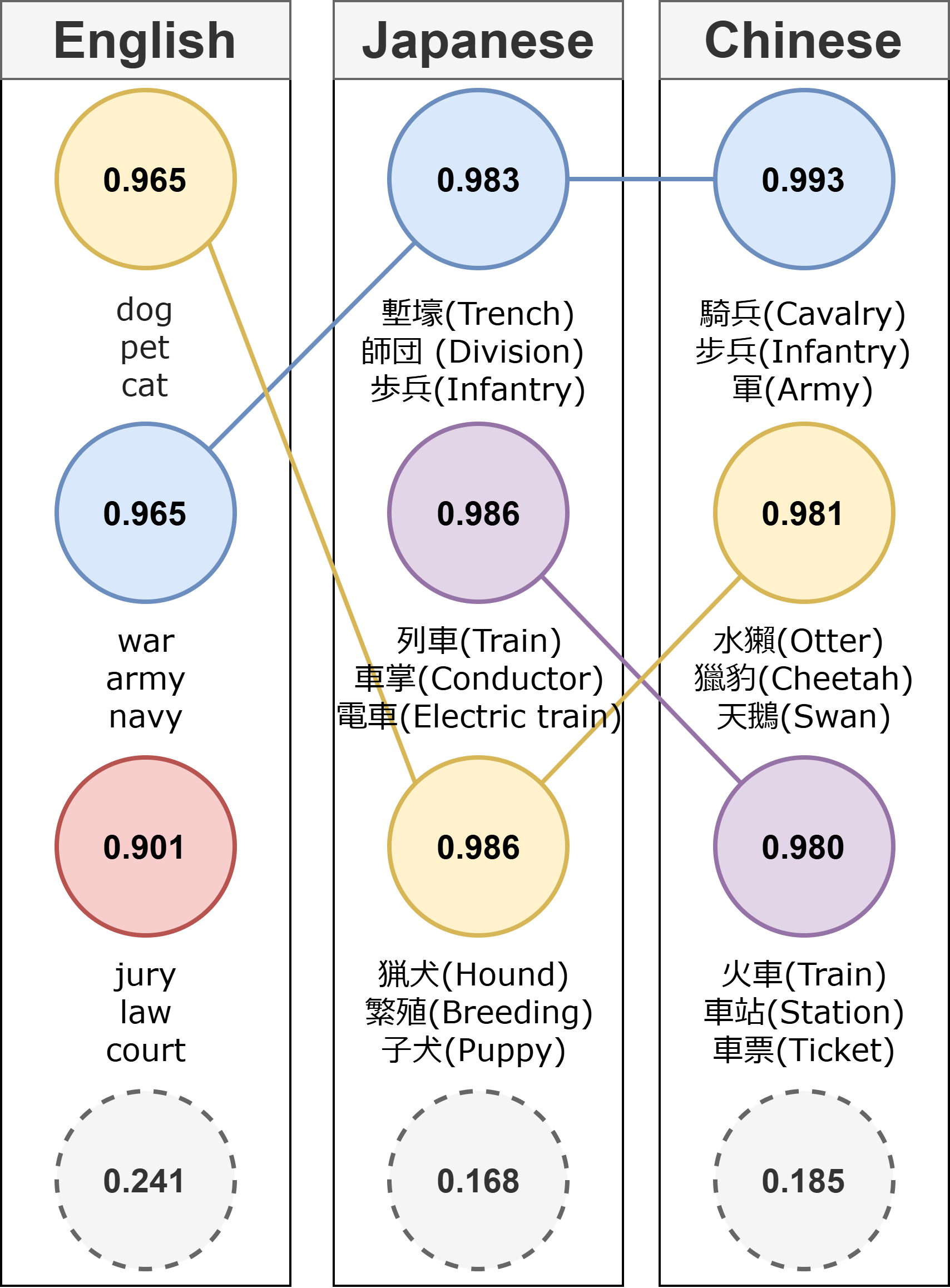}
  \caption{An illustration of clustering of independent components within and between languages. The circles represent the clusters created by Icasso, and the numbers indicate their quality indexes. Clusters with high-quality indexes were given interpretations using words. The circles connected by straight lines show components grouped together by checking consistency among languages. }
  \label{fig:illstr}
\end{figure}
\section{Introduction}
Word embedding is a technique that represents words from natural languages as multidimensional real vectors in a space (e.g., Euclidean space), making it easier to handle them as data. 
These embeddings create a continuous representation of words and sentences, facilitating data analysis and processing.
However, word embeddings are challenging to interpret because the values vary greatly depending on the training data and the dimension of the embedding space \citep{levy-goldberg-2014-dependency}.
For example, it is unclear what the embedding exactly means even if we say the embedding of ``Argentina'' is $[0.0088871, -0.02218, \dots]$. 
In order to cope with the interpretability problem, several approaches were suggested, such as Principle Component Analysis (PCA) and Independent Component Analysis (ICA, \citealp{hyevarinenbook}).

ICA gives a more interpretable representation of semantic axes (i.e., components labeled with high-relation words) over PCA \citep{musil-marecek-2024-exploring-interpretability}. 
For example, if an independent component scores high on the words ``apple'', ``banana'', and ``peach'', the semantic axis can be interpreted as the concept of fruits and be labeled as [apple banana peach]. 

\citet{yamagiwa-etal-2023-discovering} showed that ICA decomposed multilingual word embeddings into multiple interpretable axes and suggested the existence of universal semantic axes among languages. 
However, their study was limited to the calculation of correlation coefficients for the correspondence of semantic axes in each language. 
It lacked verification of the consistency of independent components within one language and across the languages. 
Consistency here refers to the reliability of independent components that appear in different runs within one language (intra-language) and the accurate correspondence of semantic axes among languages (inter-language).

In this paper, we investigate the consistency of semantic axes in both intra-language and inter-language manners as in Figure~\ref{fig:illstr}. 
First, we test each language's independent components' reliability using Icasso~\citep{HIMBERG20041214}. 
Icasso was proposed to solve the reliability problem in ICA by implementing ICA multiple times and then clustering them, enabling us to verify the consistency of each component. 
Then, after interpreting each independent component in the center of clusters (i.e., labeling each independent component with words), we correspond the semantic axes statistically. %
We apply the method proposed by \citet{hyvarinen2013testing}, originally developed for neuroscience applications, to explore common independent components across subjects. %

\section{Background}
\paragraph{ICA} 
ICA is a method to extract statistically independent components from multivariate data \citep{hyevarinenbook}. 
Let $\bm{\mathbf{X}} \in \mathbb{R}^{d \times n}$ be a data matrix, where $d$ is the data dimension and $n$ is the number of observations. %
ICA is based on the assumption that $\bm{\mathbf{X}}$ is represented as
\begin{align*}
 \bm{\mathbf{X}} = \bm{\mathbf{A}\mathbf{S}} ,
\end{align*}
where $\bm{\mathbf{A}} \in \mathbb{R}^{d \times d}$ is called the mixing matrix and $\bm{\mathbf{S}} \in \mathbb{R}^{d \times n}$ is the matrix of independent components.
Namely, the rows of $\bm{\mathbf{S}}$ correspond to $d$ latent factors that are statistically independent, and $\bm{\mathbf{A}}$ indicates how these factors are combined in each of the $d$ observed variables.
ICA employs the non-Gaussianity of independent components to compute $(\bm{\mathbf{A}},\bm{\mathbf{S}})$ from $\bm{\mathbf{X}}$.
It has been applied to various data (e.g., audio, neuroimaging data) for signal separation and feature extraction \citep{hyevarinenbook}. 

\paragraph{Interpretability in Word Embeddings}
The interpretability problem of word embeddings has been actively discussed. 
\citet{panigrahi-etal-2019-word2sense}, \citet{park-etal-2017-rotated} and \citet{bommasani-etal-2020-interpreting} aligned word embeddings in which dimensions correspond to different meanings. 
Although ICA has been a relatively new method to be utilized in interpreting word embeddings, it has shown a great explanation of semantic axes. 
\citet{musil2022independent} and \citet{musil-marecek-2024-exploring-interpretability} applied ICA to word embeddings and presented the semantic axes of the words. 
\citet{yamagiwa-etal-2023-discovering} conducted PCA and ICA on word embeddings and successfully showed that there are intrinsic semantic axes among them after comparing the results, noting ICA showed more distinctive axes. 

\section{Consistency in Embeddings}
\subsection{Intra-language Consistency}
Unlike PCA, the result of ICA may be different between different runs due to the random initialization in the algorithm, such as FastICA \citep{hyvarinen1999fast}, an insufficient number of observations, and the presence of noise in the data.
Thus, the reproducibility of the independent components needs to be verified.
\citet{HIMBERG20041214} developed a method called Icasso for assessing the algorithmic and statistical reliability of independent components. 
In this study, we apply Icasso to word embedding vectors to evaluate intrinsic semantic axes' consistency (reliability) in each language. 

Here, we briefly explain the procedure of Icasso. 
See Appendix \ref{app:def} and \citet{HIMBERG20041214} for technical details.
First, we run ICA on the data matrix $\bm{\mathbf{X}} \in \mathbb{R}^{d\times n}$ $m$ times and obtain $m$ sets of $d$ independent components. 
Then, we compute a similarity measure for each pair of two independent components from different sets (i.e., $m(m-1)/2 \cdot d^2$ pairs).
By using this similarity, we perform agglomerative hierarchical clustering of independent components.
Namely, starting from $md$ clusters of size one containing each independent component, we iteratively merge two clusters with the maximum similarity.
The reliability of each cluster is quantified by the quality index introduced in \citet{HIMBERG20041214}, which takes a value from 0 to 1.
Clusters with a quality index close to 1 represent highly reproducible independent components, which correspond to consistent semantic axes in the case of word embeddings.

\subsection{Inter-language Consistency}
While ICA can extract semantic axes for each language, it has not been quantitatively examined whether there is correspondence between the semantic axes of several languages.
Thus, we investigate the consistency of semantic axes across languages with statistical significance evaluation.
We utilize the method by \citet{hyvarinen2013testing} for clustering independent components from several data.
This method was initially developed to find common independent components across subjects in neuroimaging data analysis.

Here, we explain the method of \citet{hyvarinen2013testing} for the case of studying consistency across English and Japanese.
Suppose that we have $n$ pairs of English and Japanese words with the same meanings (e.g., ``word'' and ``単語'' (word)).
Let $\bm{\mathbf{X}}_{\mathrm{E}} \in \mathbb{R}^{d \times n}$ and $\bm{\mathbf{X}}_{\mathrm{J}} \in \mathbb{R}^{d \times n}$ be the matrices composed of their $d$-dimensional English and Japanese embedding vectors, respectively.
We apply ICA and obtain $\bm{\mathbf{X}}_{\mathrm{E}}=\bm{\mathbf{A}}_{\mathrm{E}} \bm{\mathbf{S}}_{\mathrm{E}}$ and $\bm{\mathbf{X}}_{\mathrm{J}}=\bm{\mathbf{A}}_{\mathrm{J}} \bm{\mathbf{S}}_{\mathrm{J}}$.
Recall that each row of $\bm{\mathbf{S}}_{\mathrm{E}}$ and $\bm{\mathbf{S}}_{\mathrm{J}}$ represents the activation pattern of each independent component.
Then, for $i,j=1,\dots,d$, we compute the p-value (with multiplicity correction) of the null hypothesis that the $i$-th row of $\bm{\mathbf{S}}_{\mathrm{E}}$ and the $j$-th row of $\bm{\mathbf{S}}_{\mathrm{J}}$ are independent.
If this p-value is small, the $i$-th independent component of English and the $j$-th independent component of Japanese are significantly similar.

In the above way, we compute the p-values for each pair of languages.
Then, we utilize them as a similarity measure for agglomerative hierarchical clustering of the independent components from multiple languages.
The obtained clusters indicate the consistency of semantic axes across languages.

\subsection{Interpretation of Independent Components}
\label{sec:interp}
We use three representative words selected as follows to interpret independent components obtained from word embedding vectors.
Recall that ICA is given by $\bm{\mathbf{X}}=\bm{\mathbf{A}} \bm{\mathbf{S}}$.
Thus, the embedding vector of the $j$-th word (the $j$-th column vector of \bm{\mathbf{X}}) is represented as
\begin{align*}
 \bm{\mathbf{x}}_{j} = s_{1j} \bm{\mathbf{a}}_{1} + \dots + s_{dj} \bm{\mathbf{a}}_{d},
\end{align*}
where $\bm{\mathbf{a}}_{i}$ is the $i$-th column vector of $\bm{\mathbf{A}}$ and $s_{ij}$ is the $(i,j)$-th entry of $\bm{\mathbf{S}}$.
Therefore, $s_{ij}$ quantifies how much the $j$-th word is related to the $i$-th independent component.
Based on this observation, we sort the $i$-th row of $\bm{\mathbf{S}}$ to $s_{i j_1}>s_{i j_2}>s_{i j_3}>\dots$ and take the $j_1,j_2,j_3$-th words as the representatives of the $i$-th independent component.
These words provide an intuitive understanding of the independent components as semantic axes.

\section{Experimental Settings}
We conducted the intra-language experiment focusing on the consistency of each language and then conducted the inter-language experiment focusing on the consistency among the languages. 
To align with \citet{yamagiwa-etal-2023-discovering}, we used the same FastText~\citep{joulin2016fasttext} embeddings obtained by training on 157 different languages. 
We obtained 300-dimensional embedding vectors of 50000 words for English, Japanese, and Chinese, respectively, with matrices $\bm{\mathbf{X_0}}, \bm{\mathbf{X_1}}, \bm{\mathbf{X_2}}\in\mathbb{R}^{300\times 50000}$. 
The 50000 words consist of 6903 common words among the three languages selected from the multilingual word dictionary \citep{conneau2017word} and 43097 words selected in order of their frequency of occurrence in each language by Wordfreq~\citep{wordfreq}. 
We applied Icasso implemented by \citet{captier2022biodica} to FastText's word embeddings of English, Japanese, and Chinese with 10 runs, designated 300 as the number of clusters.
We then tested the consistency among the components by the method proposed by \citet{hyvarinen2013testing} by setting the false discovery rate and the false positive rate at 1\%. Detailed explanations are in Appendix~\ref{app:FDR}.
\section{Results and Discussion}
\begin{table*}
  \centering
  \small
  \begin{tabular}{|c|c|c|}
    \hline
    \textbf{English} & \textbf{Japanese} & \textbf{Chinese} \\ \hline
 verb noun word & 流暢\ 発音\ 方言 & 話\ 流利\ 諺語 \\
           & (fluency) (pronunciation) (dialect) & (speech) (fluency) (proverb) \\ \hline
 boat sail buoy & 漁業\ 漁師\ 捕鯨 & 漁民\ 舢舨\ 捕鯨 \\
           & (fishing industry) (fisherman) (whaling) & (fisherman) (sampan) (whaling) \\ \hline
 nun pope monk & 教義\ 礼拝\ 会衆 & 恩典\ 基督\ 禱告 \\
           & (doctrine) (worship) (congregation) & (grace) (christ) (prayer) \\ \hline
 film gore cinema & 演技\ 俳優\ 演劇 & 放映\ 喜劇\ 戲服 \\
            & (acting) (actor) (drama) & (screening) (comedy) (costume) \\ \hline
 sum cosine ray & 乗法\ 整数\ 写像 & 方程\ 向量\ 切線 \\
           & (multiplication) (integer) (mapping) & (equation) (vector) (tangent) \\ \hline
 war army navy & 塹壕\ 師団\ 歩兵 & 騎兵\ 步兵\ 軍 \\
           & (trench) (division) (infantry) & (cavalry) (infantry) (army) \\ \hline
  \end{tabular}
  \caption{Interpretation of clusters.}
  \label{tab:comp_axes}
\end{table*}

\normalsize
\subsection{Overall Results}
\begin{figure}[t]
  \centering
  \includegraphics[width=\linewidth]{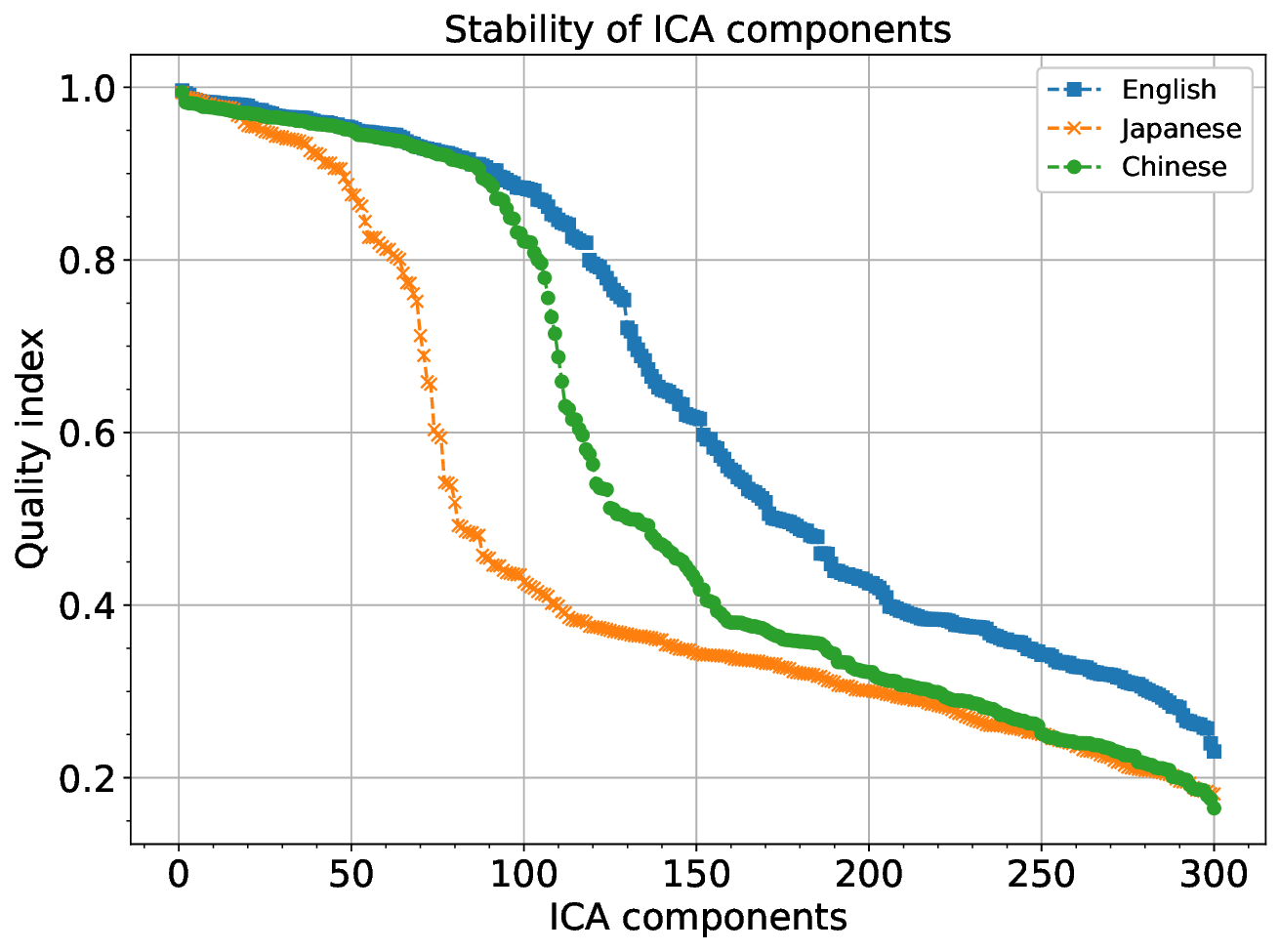}
  \caption{Quality index for FastText embeddings.}
\label{quality}
\end{figure}

Figure~\ref{quality} shows the results of Icasso. There is a clear drop after the quality index reaches 0.8. 
The number of clusters with a cluster quality index exceeding 0.8 was 118 for English, 64 for Japanese, and 104 for Chinese.

As a result of the inter-language analysis, 47 clusters, 120 out of $354\ (118\times3)$ vectors, were found, which means the average number of vectors per cluster is 2.55.
The language pairs of clusters of English-Japanese, Japanese-Chinese, Chinese-English, and all languages were 7, 10, 4, and 26, respectively. 

Based on the results of Icasso applied to static word embeddings, we identified a maximum of 118 consistent components for each language with the clusters' quality index above 0.8.
We selected the independent component located at the center of each cluster in order of the highest quality index. 
Consequently, we constructed independent component matrices \(\bm{\mathbf{S_0}}, \bm{\mathbf{S_1}}, \bm{\mathbf{S_2}} \in \mathbb{R}^{118 \times 50000}\). 

Table~\ref{tab:comp_axes} presents part of the results of the semantic axes after interpretation. 
Each axis is related to themes such as ``words'', ``fishery'', ``religion'', ``film'', ``mathematical terms'' and ``army'' demonstrating clear alignment of the axes among different languages.
\subsection{Quantitative Evaluations}
We performed human judgment experiments to evaluate the aligned components.
Five participants proficient in all three languages took part in the experiments, conducting binary classification to determine if the semantic axes were sufficiently similar. 
We tested Fleiss' kappa, which is defined in Appendix~\ref{app:def}.
$\kappa$ was 0.364, with $\bar{P}$, $\bar{P_e}$ being 0.702 and 0.531, respectively. 
This suggests that our semantic axes extracted agree fairly with human valuation since $\kappa$ is between 0.2 and 0.4 \citep{FleissKappa}.
\subsection{Discussion}
From Figure~\ref{quality}, we can see that the number of clusters formed differs among languages. 
English, as a source language in multilingual dictionary~\citep{conneau2017word}, is not surprising to have the most stable clusters. 
Japanese and Chinese were expected to have relatively the same stable clusters as target languages, but Chinese has many more clusters than Japanese. 
One possible reason for this result is that Chinese corresponds to more English words per word, more meanings are present in words, and the semantic axes are more likely to appear because of the larger sample size
\footnote{Refer to Appendix~\ref{app:dic}.}
.

However, as the results show, there are about 30\% similar semantic axes among languages, strengthening the idea of universal semantic axes' existence. 
As suggested by \citet{musil-marecek-2024-exploring-interpretability}, we can combine these robust and reliable axes to construct compositional semantic maps of word embeddings and even apply this to translation among words in different languages based on the correspondence among components.

\section{Conclusion}
Our study statistically confirmed the consistency of semantic axes within and across languages using ICA components.
Recognizing the inherent instability of ICA, we employed Icasso to ensure robustness by running multiple iterations and clustering the results. 
This process resulted in high-quality, reproducible semantic axes for English, Japanese, and Chinese. 
We then statistically verified inter-language consistency by identifying common semantic axes shared among these languages, supported by rigorous statistical tests. 
Our primary contribution is the innovative use of statistical methods to ensure the reliability and universality of semantic axes. 
The validation underscores the effectiveness of our approach in achieving consistent and interpretable word embeddings and highlights the potential for improved multilingual natural language processing applications.

\section*{Limitations}
We also conducted experiments with BERT because it would have been beneficial to include an analysis of contextualized word embeddings compared to static word embeddings like FastText.
It would be necessary to have a parallel corpus across English, Japanese, and Chinese to gain word embeddings from the same context. 
However, there currently needs to be more data on the multilingual parallel corpus, especially in English, Japanese, and Chinese. 
For preliminary experiments, we used TED Multilingual Parallel Corpus\footnote{\url{https://github.com/ajinkyakulkarni14/TED-Multilingual-Parallel-Corpus}}. 
However, ICA did not converge, which was mainly attributed to the low amount of data, so we did not include the results in this paper.

Also, the number of independent components was limited to the dimensionality because the linear trait of ICA. 
Non-linear ICA proposed by \citet{hyvarinen2019nonlinear} were not implemented due to time constraints but can be applied to word embeddings in the future.

\section*{Acknowledgements}
This work was partially supported by PRESTO, JST Grant Number JPMJPR21C8, Japan, and JSPS KAKENHI Grant Number 22K17865.
\bibliography{anthology,custom}
\newpage
\appendix

\section{Definitions}
\label{app:def}
\subsection{Similarity}
The similarity \( \sigma_{ij} \) between \( s_i \) and \( s_j \) is defined as follows:
\[
\sigma_{ij}=\left|\frac{\frac{1}{d}\sum_{k}s_{ik}s_{jk}}{\sqrt{\frac{1}{d}\sum_{k}s_{ik}^2} \sqrt{\frac{1}{d}\sum_{k}s_{jk}^2}}\right|
\]
In other words, \( \sigma \) is the absolute value of the correlation coefficient, and the degree of difference is given by
\[
d_{ij}=1-\sigma_{ij}.
\]
\subsection{Quality Index}
The quality index \( I_q \) is defined as follows:
\begin{align*}
 I_q(C_m) = \frac{1}{|C_m|^2} &\sum_{i,j \in C_m} \sigma_{ij} \\ 
&-\frac{1}{|C_m||C_{-m}|} \sum_{i \in C_m} \sum_{j \in C_{-m}} \sigma_{ij},
\end{align*}
where \( C_m \) refers to cluster \( m \) and \( C_{-m} \) refers to all independent components except cluster \( m \). \( |C_m| \) is the number of components in a cluster.

\subsection{Fleiss' Kappa}
Fleiss' kappa is defined as below:
\begin{align*}
  \kappa=\frac{\bar{P}-\bar{P_e}}{1-\bar{P_e}}.
\end{align*} 
\section{FDR and FPR}
\label{app:FDR}
In multiple inter-language tests, the null hypothesis may be rejected by chance as the number of tests increases. 
For example, if we consider the test at 5\% significance level, when all the null hypotheses are true, as many as 50 null hypotheses are rejected by chance in 1000 tests. 
Therefore, as discussed below, a correction is often made to account for this.
The false discovery rate (FDR) is defined as
\[
\text{FDR} = \frac{
 \text{False rejections when } H_0 \text{ is true}
}{
 \text{Total rejections}
},
\]
where $H_0$ is the null hypothesis.
To keep the FDR below a specified value of $\alpha_{FD}$ in overall tests, the corrected FDR $\alpha^{corr}_{FD}$ is calculated in each test by the method proposed by \citet{FDR}.
In addition to FDR, we also consider the false positive rate (FPR). The FPR is defined as follows:
\[
\text{FPR} = \frac{
 \text{False rejections when } H_0 \text{ is true}
}{
 \text{Cases where } H_0 \text{ is false}
}.
\]
To control FDR below $\alpha_{FP}$, corrected value $\alpha^{corr}_{FP}$ is also calculated by Bonferroni correction \citep{hyvarinen2011testing}. 
In the experiment, FPR was used to confirm the existence of clusters among languages, and FDR was used to decide which components should be clustered into existing clusters.

\section{Detailed Results}
\begin{figure}[tbh]
    \includegraphics[width=\linewidth]{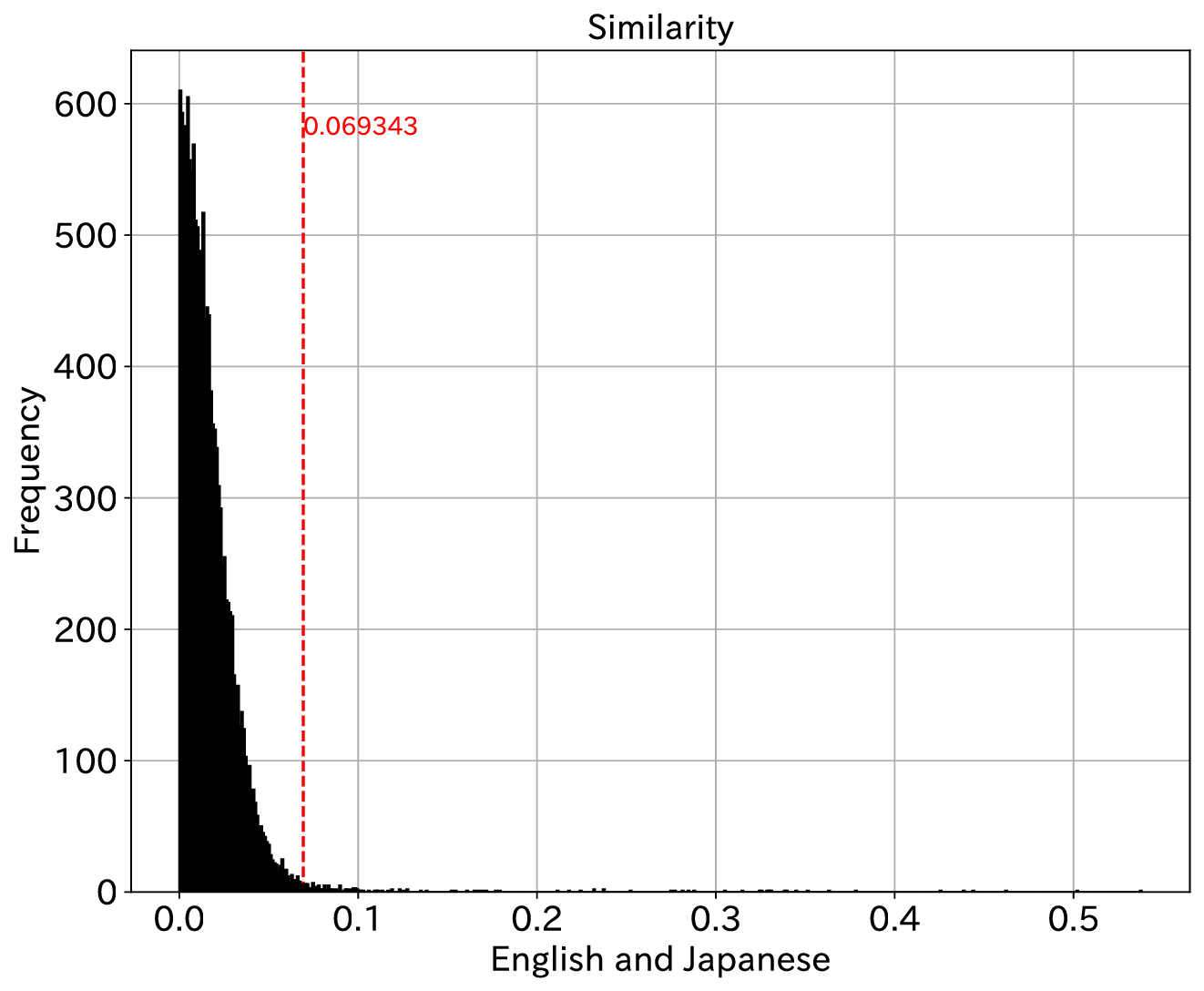}
    \caption{Similarity of Independent Components - English and Japanese.}
    \label{fig:simEJ}
  \end{figure}

\begin{figure}[tbh]
    \includegraphics[width=\linewidth]{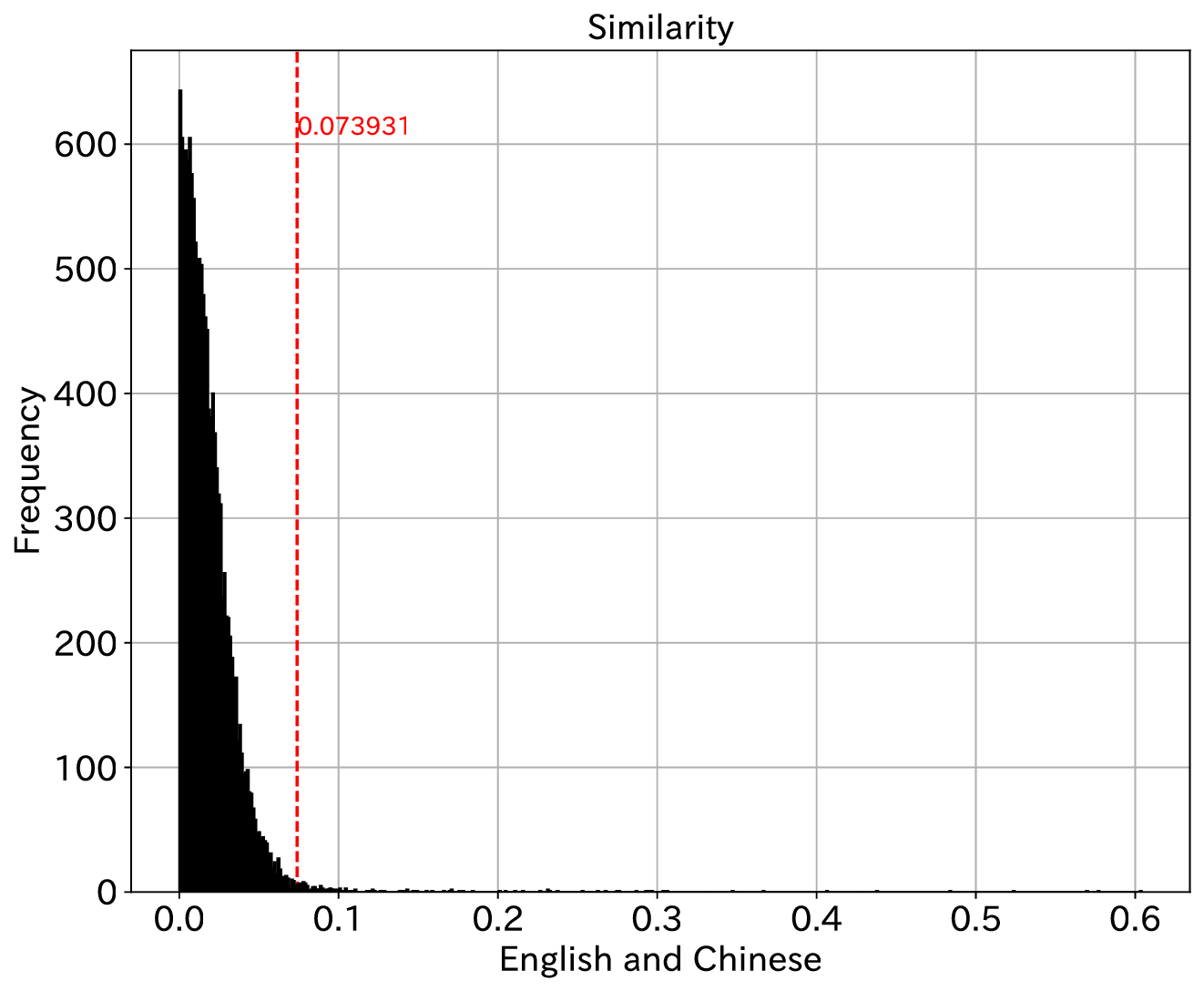}
    \caption{Similarity of Independent Components - English and Chinese.}
    \label{fig:simEC}
  \end{figure}

\begin{figure}[tbh]
    \includegraphics[width=\linewidth]{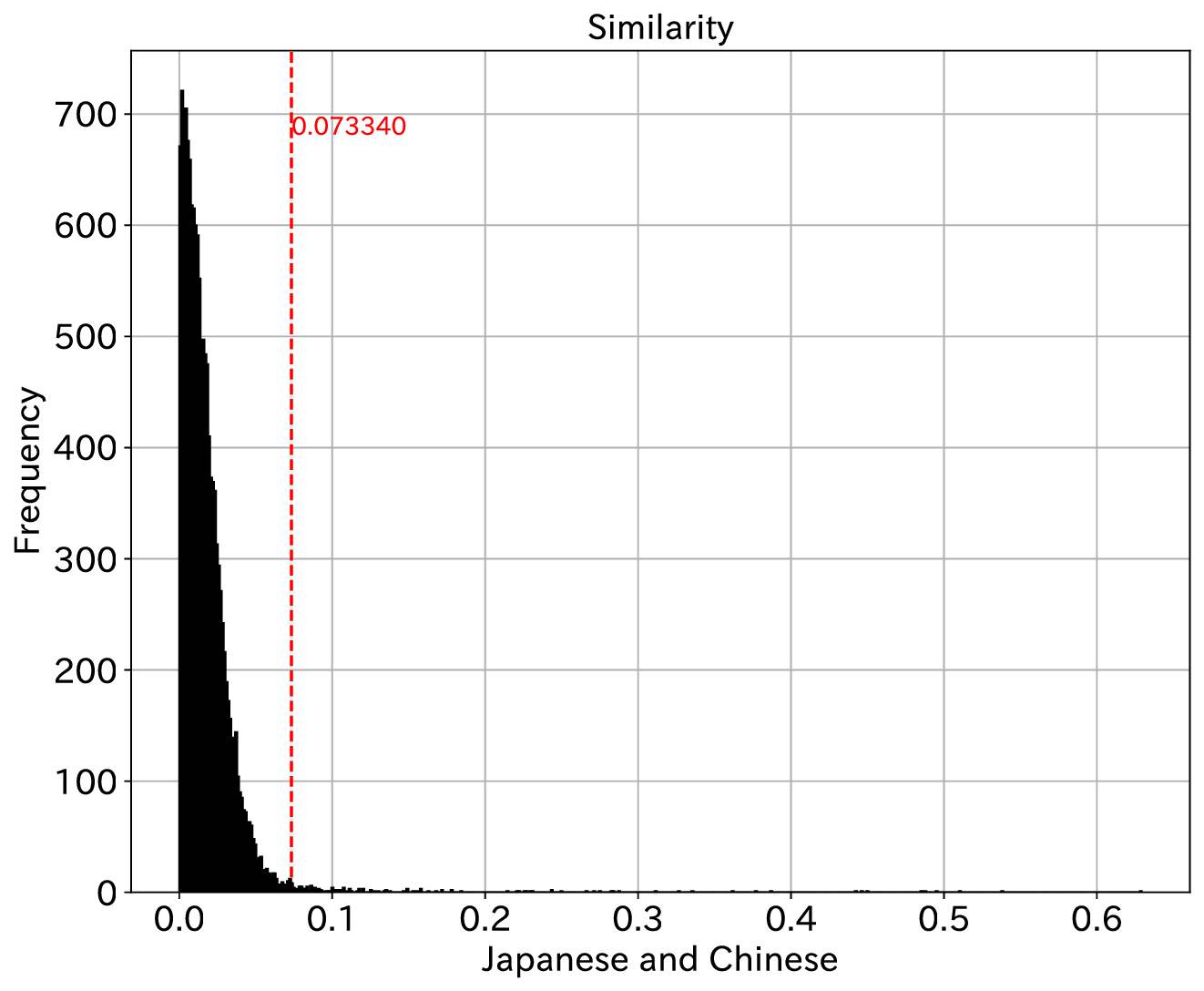}
    \caption{Similarity of Independent Components - Japanese and Chinese.}
    \label{fig:simJC}
  \end{figure}

Table~\ref{tab:Icasso} shows detailed results of ICA. 
The distribution of similarities is illustrated in Figure~\ref{fig:simEJ}, Figure~\ref{fig:simEC}, and Figure~\ref{fig:simJC}.
The red lines in the figures represent the top 5\% line of similarities. 
\begin{table*}
  \centering
  \begin{tabular}{|l|l|}
    \hline
 Number of Clusters Found & 47 \\
 Number of Clustered Vectors & 120 (33.90\% of all vectors) \\
 Average Number of Vectors per Cluster & 2.55 \\
    \hline
    \multicolumn{2}{|c|}{Internal Parameters} \\
    \hline
    $\alpha_{FD}^{corr}$ & $1.000000 \times 10^{-2}$ \\
 Minimum Similarity Considered Significant by FDR & 0.1110 \\
    $\alpha_{FP}^{corr}$ & $2.754821 \times 10^{-5}$ \\
 Minimum Similarity Considered Significant by FPR & 0.1468 \\
    \hline
  \end{tabular}
  \caption{Detailed results of ICA.}
  \label{tab:Icasso}
\end{table*}
\section{Dictionary Statistics}
The number of unique Japanese words in the English-Japanese dictionary was 21003, and the number of English words was 22531.
The number of unique Chinese words in the English-Chinese dictionary was 13768, and the number of English words was 25969. 
\label{app:dic}
\section{Questionnaire Form}
The following questionnaire form~\ref{mybox}, initially in Japanese, was used to conduct quantitative evaluations of semantic axes. 
The English translations of Japanese and Chinese words are only for explanation here and were not annotated in the actual questionnaire form.
\newpage
\newpage
\begin{tcolorbox}[title=Questionnaire Form, label=mybox]
Below is a list of words in several languages. If you think that the English, Japanese, and Chinese words all belong to the same meaning category, check the box.
For example,

en:[`eyes' `see' `rib'] ja:[`視界'(`vision') `網膜'(`retina') `凝視'(`stare')] zh:[`觀看'(`look') `凝視'(`stare') `眼'(`eye')]

In this case, the three languages have a meaning associated with the eye, so check the box.

en:[`deco' `arts' `murals' `dali' `vase'] ja:[`礼儀'(`courtesy') `ひも'(`string') `冗長'(`redundancy')] zh:[`民俗'(`folk') `漆器'(`lacquerware') `壁畫'(`wall art')]

In this case, because the list of Japanese words does not make sense or does not match the meaning of the other languages, do not check the box.
\end{tcolorbox}

\end{document}